\documentclass{article}

 \usepackage[preprint]{neurips_2025}
\usepackage{graphicx}

\usepackage[utf8]{inputenc} % allow utf-8 input 
\usepackage[T1]{fontenc}    % use 8-bit T1 fonts
\usepackage{hyperref}       % hyperlinks
\usepackage{url}            % simple URL typesetting
\usepackage{booktabs}       % professional-quality tables
\usepackage{amsfonts}       % blackboard math symbols
\usepackage{nicefrac}       % compact symbols for 1/2, etc.
\usepackage{microtype}      % microtypography
\usepackage{xcolor}         % colors

\title{BehaviorVLM: Unified Finetuning-Free Behavioral Understanding with Vision-Language Reasoning}

\author{%
  Jingyang Ke\thanks{Equal contribution.} \\
  Georgia Institute of Technology \\
  Atlanta, GA, 30332 \\
  \texttt{jingyang.ke@gatech.edu} \\
  \And
  Weihan Li\footnotemark[1] \\
  Georgia Institute of Technology \\
  Atlanta, GA, 30332 \\
  \texttt{weihanli@gatech.edu} \\
  \And
  Amartya Pradhan \\
  Georgia Institute of Technology, Emory University \\
  Atlanta, GA, 30322 \\
  \texttt{amartya.pradhan@emory.edu} \\
  \And
  Jeffrey E. Markowitz \\
 Georgia Institute of Technology, Emory University \\
  Atlanta, GA, 30332 \\
  \texttt{jeffrey.markowitz@bme.gatech.edu} \\
  \And
  Anqi Wu \\
  Georgia Institute of Technology \\
  Atlanta, GA, 30332 \\
  \texttt{anqiwu@gatech.edu} \\
}

\begin{document}

\maketitle

\begin{abstract}
Understanding freely moving animal behavior is central to neuroscience, where pose estimation and behavioral understanding form the foundation for linking neural activity to natural actions.
Yet both tasks still depend heavily on human annotation or unstable unsupervised pipelines, limiting scalability and reproducibility.
We present \textbf{BehaviorVLM}, a unified vision-language framework for pose estimation and behavioral understanding that requires \textbf{no task-specific finetuning} and \textbf{minimal human labeling} by guiding pretrained Vision-Language Models (VLMs) through detailed, explicit, and verifiable reasoning steps.
For pose estimation, we leverage quantum-dot-grounded behavioral data and propose a multi-stage pipeline that integrates temporal, spatial, and cross-view reasoning.
This design greatly reduces human annotation effort, exposes low-confidence labels through geometric checks such as reprojection error, and produces labels that can later be filtered, corrected, or used to fine-tune downstream pose models.
For behavioral understanding, we propose a pipeline that integrates deep embedded clustering for over-segmented behavior discovery, VLM-based per-clip video captioning, and LLM-based reasoning to merge and semantically label behavioral segments.
The behavioral pipeline can operate directly from visual information and does not require keypoints to segment behavior.
Together, these components enable scalable, interpretable, and label-light analysis of multi-animal behavior.
\end{abstract}

\section{Introduction}
\label{sec:intro}

Understanding freely moving animal behavior is central to neuroscience.
Two fundamental tasks are pose estimation and behavioral segmentation, which together provide the bridge between neural activity and natural action.
In practice, however, both problems still require substantial human labor.
Pose estimation toolkits such as DeepLabCut~\citep{mathis2018deeplabcut}, SLEAP~\citep{pereira2022sleap}, and Lightning Pose~\citep{biderman2024lightning} can achieve strong accuracy, but each new experimental setup usually requires manual labels before training can begin.
Pretrained foundation models such as SuperAnimal~\citep{ye2024superanimal} reduce this burden, yet they still depend on human-labeled pretraining data and can degrade under new camera geometries, imaging conditions, or animal morphologies.
Behavioral understanding faces a parallel limitation.
In this paper, \textit{behavioral understanding} refers to behavioral segmentation together with a human-understandable interpretation for each segment.
Recent VLM- and LLM-based systems such as MouseGPT~\citep{xu2025mousegpt} and AmadeusGPT~\citep{ye2023amadeusgpt} show that language models can help describe animal behavior, but they do not replace the full annotation workflow that a human analyst performs when identifying transitions and assigning semantic labels to behavior segments.
At the other extreme, unsupervised approaches such as MotionMapper \citep{berman2014mapping}, MoSeq \citep{wiltschko2015mapping}, and Keypoint-MoSeq~\citep{weinreb2024keypoint} scale well, but they often produce segments that are difficult to interpret, switch too rapidly, or do not align cleanly with human-understandable behavioral categories.
This limitation arises because these methods typically rely on keypoints or low-dimensional motion representations and do not directly extract semantic labels from the visual evidence in the video.

\begin{figure}[t]
  \centering
  \includegraphics[width=\linewidth]{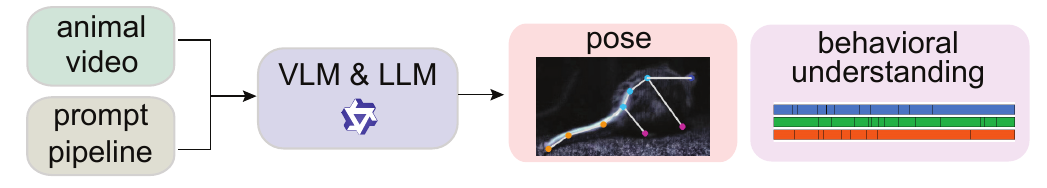}
  \caption{\textbf{Overview of BehaviorVLM.}
  This VLM \& LLM-based framework addresses pose estimation and behavioral understanding with minimum manual labeling and no finetuning.}
  \label{fig:behaviorVLM}
\end{figure}

We present \textbf{BehaviorVLM}, a unified vision-language framework that addresses both pose estimation and behavioral understanding \textbf{without task-specific finetuning} and with \textbf{minimal human labeling}, by guiding pretrained Vision-Language Models (VLMs) through structured, multi-stage reasoning pipelines.
The central idea is to mimic how a human would carry out these annotation tasks in practice.
Rather than asking a model for a final answer in a single step, we decompose each task into explicit intermediate stages that use visual evidence, expose uncertainty, and allow labels to be reviewed or corrected afterward.
This framing is especially useful when the goal is to replace large amounts of manual work rather than to claim that every automatic label is perfect.

For \textbf{pose estimation}, we leverage near-infrared fluorescent quantum dots (QDs) \citep{QD} injected at body keypoints to provide candidate keypoint locations across six synchronized camera views.
A VLM is guided through a multi-stage reasoning pipeline that integrates temporal, spatial, and cross-view constraints to predict accurate 3D keypoint trajectories.
The pipeline requires only three manually labeled seed frames, and completed predictions are appended to a rolling window and reused as few-shot exemplars for later frames.
The QD signals substantially reduce human labeling effort compared with conventional manual annotation, reduce bias from imprecise human labels, and make it possible to identify poor pseudo-labels after the fact using geometric criteria such as large 3D reprojection error.
Those filtered labels can then be used directly or used to fine-tune a downstream pose estimation model.
More broadly, this setup encourages the use of QD-based labeling for small animals such as mice, fish, and birds, where conventional pose annotation or motion tracking with motion capture devices is especially difficult.

For \textbf{behavioral understanding}, we introduce a multi-stage pipeline that first applies deep embedded clustering to obtain fine-grained, over-segmented behavioral clips for each animal, then invokes a VLM to generate per-clip behavioral labels and natural-language descriptions, and finally leverages an LLM to merge similar segments and assign semantically meaningful labels.
This pipeline makes heavy use of visual information.
In particular, the segmentation process can operate directly on video features and does not require keypoints, which distinguishes it from prior behavior pipelines that are restricted to pose-based inputs.

Together, these two pipelines form a unified framework (Figure~\ref{fig:behaviorVLM}) that replaces extensive human annotation and task-specific model training with structured vision-language reasoning, enabling scalable and interpretable automated analysis of naturalistic animal behavior.
Our main contributions are:
\begin{itemize}
    \item A multi-stage VLM-based reasoning pipeline for QD-grounded pose estimation that requires only three labeled seed frames and produces labels that can be inspected, filtered, corrected, and reused for downstream pose model fine-tuning.
    \item A multi-stage behavioral understanding pipeline that converts visual or fused behavioral features into semantically meaningful behavioral segments through low-cost over-segmentation, VLM-based visual interpretation, and LLM-based semantic reasoning.
    \item Evaluation on a custom six-view quantum-dot mouse dataset~\citep{QD} and the MABe2022 Mouse Triplets benchmark~\citep{sun2023mabe22}, demonstrating that finetuning-free vision-language reasoning can achieve reliable pose estimation and interpretable multi-animal behavioral segmentation.
\end{itemize}

\section{Pose Estimation}
\subsection{Experimental Data}
\begin{figure}[t]
  \centering
  \includegraphics[width=\linewidth]{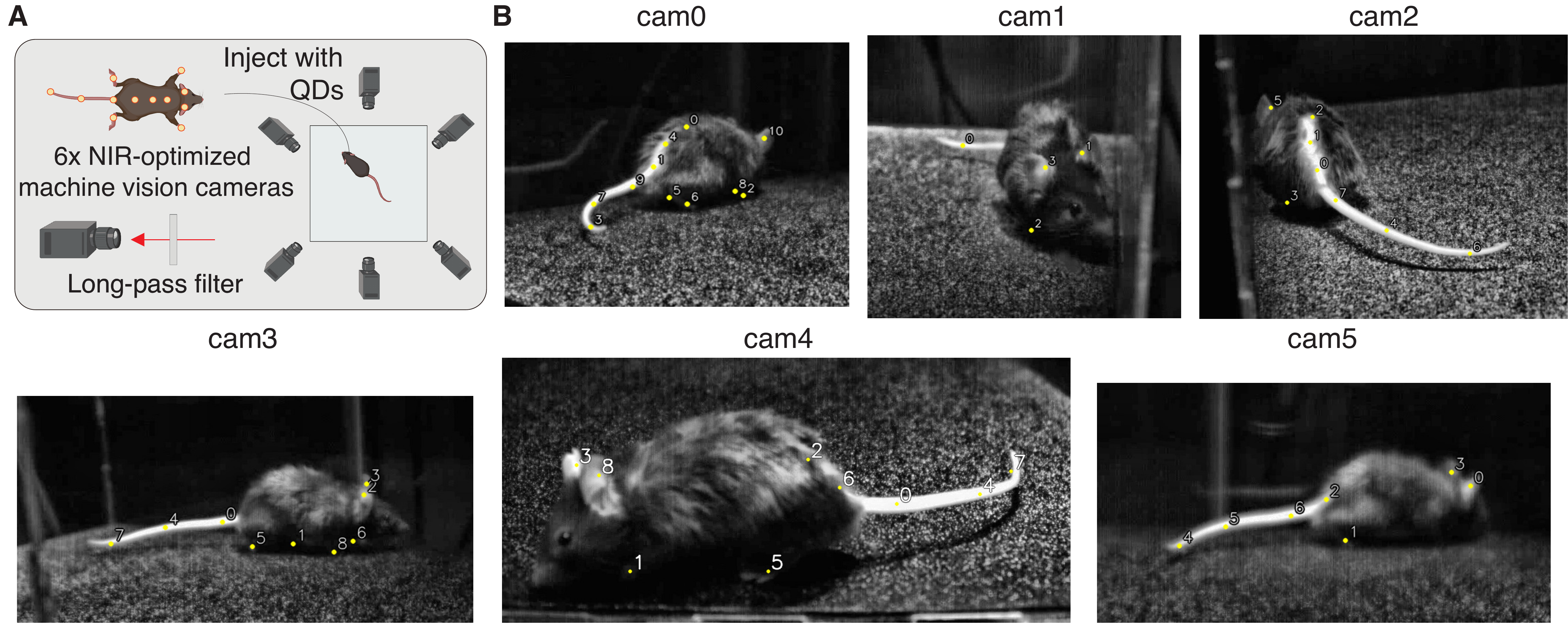}
  \caption{\textbf{Pose estimation experimental setup.}
  \textbf{(A)}~Data collection: a mouse injected with near-infrared quantum dots (QDs) at 12 body keypoints is recorded by six synchronized NIR-optimized cameras
  \textbf{(B)}~Example six-view frames with QD fluorescence centroids detected and overlaid as numbered candidates on the reflectance images. Centroid indices are local to each view; the goal is to assign anatomical identities to these candidates across all cameras and timepoints.}
  \label{fig:qd}
\end{figure}

We use a dataset of 500 synchronized timepoints from six cameras recording a freely moving mouse.
The mouse was injected with near-infrared fluorescent nanoparticles (quantum dots, QDs) at 12 anatomical keypoints, following the QD data acquisition procedure in~\citep{QD}.
This setup provides both reflectance images, which capture the visible behavior of the animal, and fluorescence images, which reveal the QD signals at body locations (Figure~\ref{fig:qd}A).
Each fluorescence centroid indicates the location of a body marker, but not its anatomical identity.
The raw frames have resolution $2048\times 1400$ in every view.

For each frame and camera, we apply Segment Anything 3~\citep{SAM3} to detect the mouse body mask.
We then crop the frame to the mask's tight bounding box and pad the crop by 16 pixels plus 5\% of the bounding box dimensions.
Because the apparent size and position of the mouse vary across viewpoints and over time, the crop dimensions differ across cameras and frames.
QD centroid locations are extracted from the fluorescence channel inside each crop and overlaid as numbered candidate keypoints on the reflectance image (Figure~\ref{fig:qd}B).

The task is to assign anatomical identities to these candidate points with minimal human effort.
We use only three manually labeled seed frames.
Afterward, the pipeline generates labels automatically, and these labels can be reviewed, corrected, or filtered using geometric confidence measures such as large 3D reprojection error before they are used in downstream analysis or pose model fine-tuning.

\subsection{Method}
\label{sec:pose_method}

\begin{figure}[t]
  \centering
  \includegraphics[width=\linewidth]{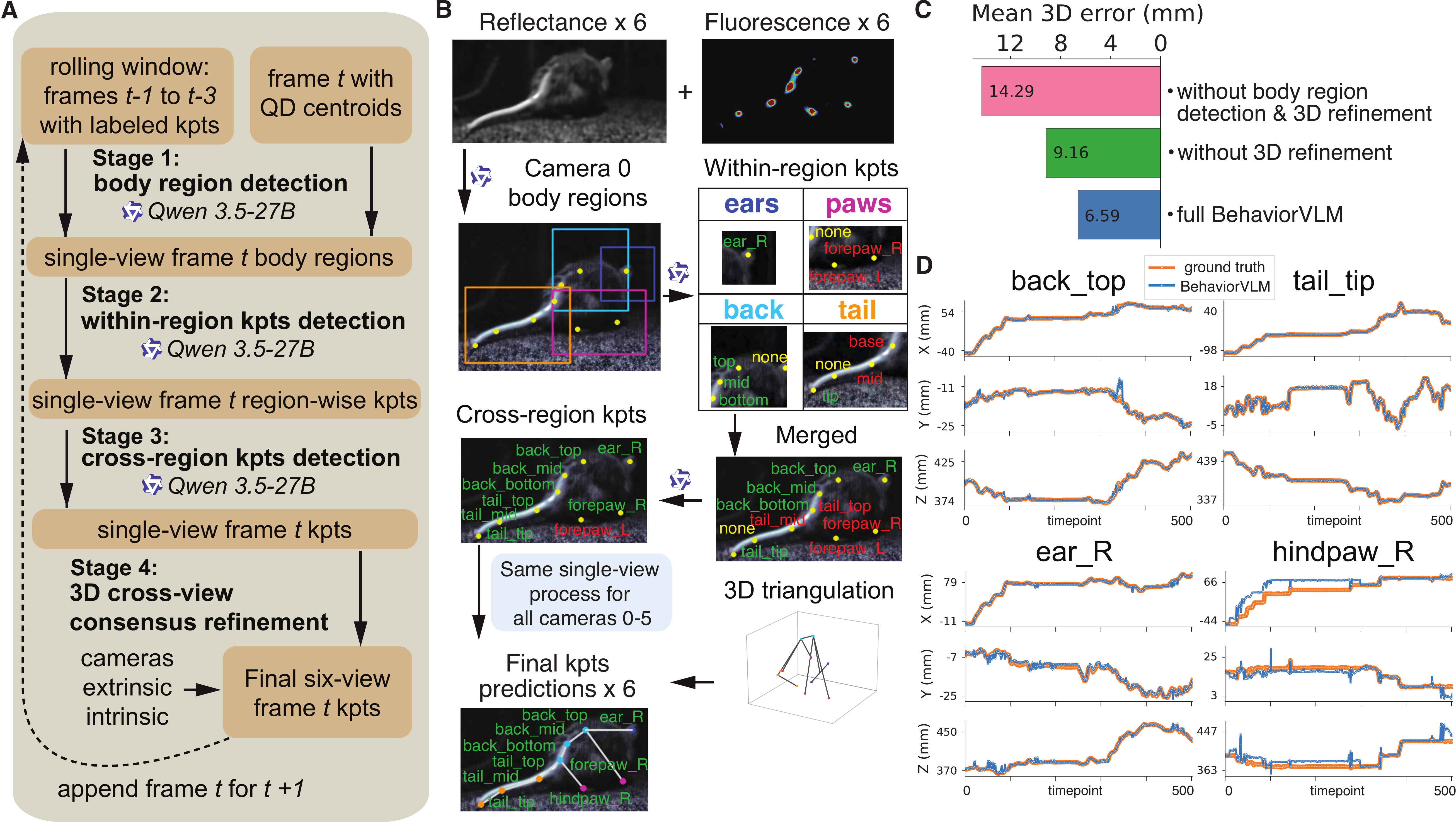}
  \caption{\textbf{BehaviorVLM pose estimation pipeline and results.}
  \textbf{(A)}~pipeline overview.
  \textbf{(B)}~Detailed example for one frame from camera~0: the VLM first localizes four body regions (ears, back, paws, tail) via bounding boxes, then assigns centroids to keypoints within each cropped region, merges assignments, and resolves conflicts. Six-view predictions are triangulated into 3D and refined via RANSAC consensus.
  \textbf{(C)}~Ablation study showing mean 3D keypoint error (mm) averaged over 12 keypoints across 500 frames. The full BehaviorVLM pipeline (6.59\,mm) outperforms variants without 3D cross-view refinement (9.16\,mm) and without both body region detection and 3D refinement (14.29\,mm), demonstrating the contribution of each component.
  \textbf{(D)}~Representative 3D keypoint trajectories for four body keypoints: \texttt{back\_top}, \texttt{tail\_tip}, \texttt{ear\_R}, and \texttt{hindpaw\_R}. Ground truth is shown in orange, BehaviorVLM predictions in blue.}
  \label{fig:pose}
\end{figure}

BehaviorVLM formulates QD-grounded pose estimation as a structured visual reasoning problem, guiding a vision-language model (VLM) through a four-stage pipeline (Figure~\ref{fig:pose}A,\,B).
The pipeline requires only three manually labeled seed frames.
Completed predictions are appended to a rolling window and reused as few-shot exemplars for subsequent frames, enabling temporally coherent keypoint tracking.

\paragraph{Stage 1: Body Region Detection.}
The 12 body keypoints are partitioned into four anatomical regions: \textit{ears} (\texttt{ear\_L}, \texttt{ear\_R}), \textit{back} (\texttt{back\_top}, \texttt{back\_middle}, \texttt{back\_bottom}), \textit{paws} (\texttt{forepaw\_L}, \texttt{forepaw\_R}, \texttt{hindpaw\_L}, \texttt{hindpaw\_R}), and \textit{tail} (\texttt{tail\_base}, \texttt{tail\_middle}, \texttt{tail\_tip}).
For each region in each camera view, the VLM (Qwen~3.5-27B~\citep{qwen3.5}) is provided with three consecutive preceding frames as rolling few-shot exemplars, each annotated with a colored bounding box over the target region.
The VLM predicts the bounding box of that region in the current frame.
This stage narrows the search space before any keypoint identity is assigned and helps the pipeline remain stable during fast motion and partial occlusion.

\paragraph{Stage 2: Within-Region Keypoint Assignment.}
The target frame is cropped to each predicted region bounding box.
The VLM is then prompted with three rolling exemplar crops, each with verified centroid-to-keypoint assignments, and asked to assign the numbered centroids inside the crop to the corresponding region keypoints.
This decomposition into local crops reduces assignment ambiguity because each crop contains only 2--4 relevant keypoints.

\paragraph{Stage 3: Cross-Region Assignment Reconciliation.}
Per-region assignments from Stage~2 are merged across all four regions into a single full-frame assignment.
At this stage, some conflicts can remain, such as two keypoints being assigned to the same centroid or some visible centroids being left unused.
We therefore call the VLM once more with the full-frame image and a structured description of the current partial assignments and candidate centroid indices.
The VLM reconciles conflicts and fills gaps so that the visible centroids receive a complete and unique assignment.

\paragraph{Stage 4: 3D Cross-View Consensus Refinement.}
Given the per-camera 2D keypoint predictions across all six views, we apply a RANSAC-based triangulation \citep{fischler1981random} and cross-view consistency correction to refine potentially erroneous centroid assignments.
For each keypoint, we first triangulate a 3D world position using RANSAC over subsets of cameras.
We select the subset that maximizes the inlier count, where inliers are cameras whose reprojection error falls below threshold $\tau_{\mathrm{reproj}}$, and then re-triangulate using only those inlier cameras to obtain a refined 3D estimate.
We next compute the reprojection error of this 3D estimate in every camera and partition assignments into \emph{locked} cameras (low error, trusted) and \emph{target} cameras (high error, to be corrected).
For each high-error keypoint, we enumerate hypotheses by considering alternative nearby centroid assignments in each target camera and projecting the current 3D estimate to identify geometrically plausible candidates.
Each hypothesis is scored by re-running RANSAC triangulation and computing the resulting mean reprojection error.
We accept the hypothesis with the lowest error and resolve conflicting assignments through swaps when necessary.

This final stage matters not only for accuracy, but also for quality control.
The same reprojection-based confidence measure can be used after prediction to identify low-quality labels, remove them, or send them for manual correction before training a downstream pose estimation model.
The completed frame-$t$ predictions are then appended to the rolling window and used as exemplars for frame $t+1$.

\subsection{Results}
\label{sec:pose_results}

\paragraph{Quantitative Evaluation.}
Figure~\ref{fig:pose}C reports the mean 3D keypoint prediction error across all 12 keypoints over the 500-timepoint recording.
To evaluate the contribution of each pipeline component, we compared three versions of BehaviorVLM: (i) the \textbf{full BehaviorVLM pipeline}, (ii) \textbf{BehaviorVLM without 3D cross-view refinement}, and (iii) \textbf{BehaviorVLM without region detection \& 3D refinement} (plain rolling three-shot prompting without region-based decomposition or 3D cross-view refinement).
Both the region-based decomposition and the 3D cross-view refinement contribute substantially to accuracy, with the full pipeline reducing mean error by 54\% relative to the na\"ive baseline.

\paragraph{Qualitative Evaluation.}
Figure~\ref{fig:pose}D shows representative predicted keypoint trajectories for each of the four body regions (back, tail, ears, paws).
For back, tail, and ear keypoints, BehaviorVLM tracks the trajectories closely across the full recording.
When predictions temporarily deviate from ground truth, the pipeline often recovers in later frames instead of drifting through the rest of the sequence.
This resilience is important in practice.
Even when a few frames are labeled imperfectly and those labels are reused as exemplars, the VLM does not simply copy the earlier mistake.
Its visual reasoning still allows a later frame to be judged somewhat independently, which helps the system correct earlier errors rather than accumulate them monotonically over time.
Paw keypoints remain the hardest case because of frequent occlusion and strong visual similarity between left and right limbs and between forepaws and hindpaws.
BehaviorVLM still occasionally confuses these identities.
These errors can be identified later using the same geometric confidence checks from Stage~4 and then corrected manually, removed, or used selectively when constructing downstream training data.
Overall, the results show that BehaviorVLM can generate useful and reviewable pose labels from QD-grounded videos using only three labeled seed frames and no task-specific fine-tuning.

%-----------------------------------------------------------------------
\section{Behavioral Understanding}
\label{sec:behavior}

\subsection{Experimental Data}
\label{sec:behavior_exp}

We evaluate the behavioral understanding pipeline on the Mouse Triplets dataset from the MABe2022 challenge~\citep{sun2023mabe22}, which consists of top-view videos of three freely interacting mice in an open arena equipped with a food zone.
Each video is annotated with frame-level behavior labels that include \textit{chase}, \textit{huddle}, \textit{oral contact}, and \textit{oral-genital contact}.
These labels are human annotations provided by the dataset.
In the experiments reported here, however, we use only the videos and derived visual features as inputs to our pipeline and do not use these manual labels during segmentation or semantic interpretation.

\subsection{Method}
\label{sec:behavior_method}

\begin{figure}[t]
  \centering
  \includegraphics[width=0.85\linewidth]{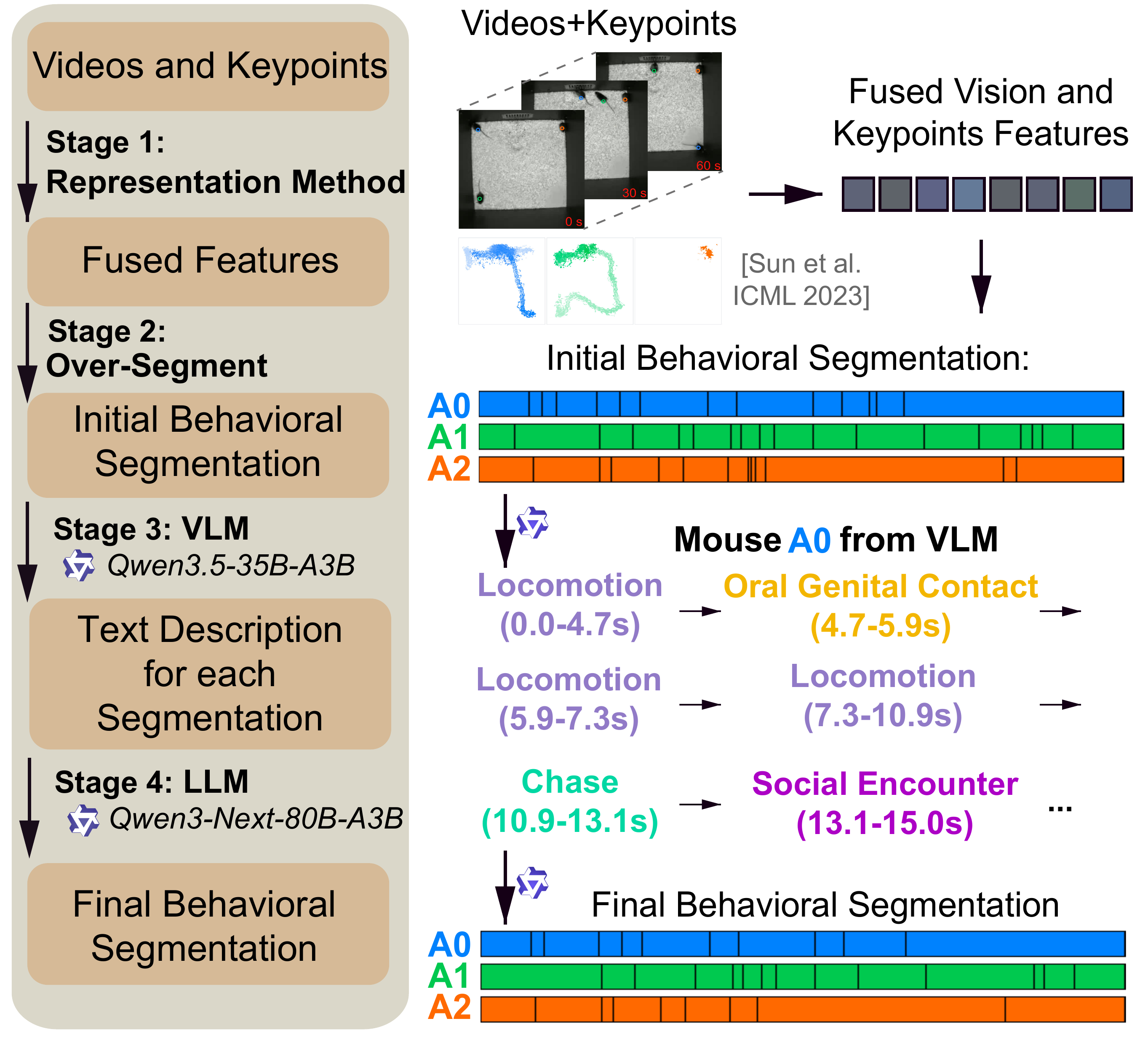}
  \caption{Overview of the BehaviorVLM pipeline for semantic behavioral understanding. Behavioral features are first over-segmented into fine-grained candidate clips for each animal. A vision-language model (VLM) then generates natural-language labels and descriptions for each clip. The mouse \texttt{A0} segments shown here are the direct VLM-stage segments and are therefore more fine-grained than the final LLM-merged mouse \texttt{A0} segments shown in Figure~\ref{fig:behavior_results}.}
  \label{fig:behavior_pipeline}
\end{figure}

BehaviorVLM provides a pipeline for semantic behavior segmentation in multi-animal videos (Figure~\ref{fig:behavior_pipeline}).
Given behavioral feature representations, the method converts low-level temporal structure into interpretable behavioral segments.
The VLM first interprets what happens in each short clip.
The LLM then merges neighboring clips into temporally coherent semantic descriptions of individual and social behaviors.
This is a human-like process: observe actions, describe them, and then merge them into meaningful behaviors.

\paragraph{Stage 1: Flexible Feature Representation.}
BehaviorVLM operates on behavioral feature representations extracted from multi-animal videos.
In our implementation, we use fused visual and keypoint features produced by the LookAgain framework~\citep{li2026learninglookondemandkeypointvideo}, which integrates visual appearance and motion information from keypoints into a unified representation.
More generally, the pipeline can accept different types of behavioral features, including: (i) \textit{keypoint-based features}, derived from tracked body keypoints (e.g., pairwise distances, angles, velocities) or pretrained motion encoders; (ii) \textit{visual features}, extracted directly from raw video frames using a pretrained visual encoder; or (iii) \textit{fused features}, combining both keypoint and visual streams.
This flexibility allows BehaviorVLM to operate when only partial modalities are available, supports simultaneous analysis of multiple animals, and makes the method more robust to keypoint noise, missing keypoints, and changes in body orientation or camera rotation.

\paragraph{Stage 2: Over-Segmented Behavior Discovery via Deep Embedded Clustering.}
Given the behavioral feature representations, we apply Deep Embedded Clustering (DEC)~\citep{xie2016unsupervised} to discover initial behavioral segments.
During training, DEC is optimized jointly across all animals by minimizing the sum of per-animal clustering losses.
At inference time, the learned clustering model is applied to each animal's feature sequence separately to produce behavioral segmentation for that animal.
We intentionally use a relatively large number of clusters to produce short, fine-grained video clips, where each clip corresponds to a contiguous behavioral segment of a single animal with relatively homogeneous motion statistics.

This over-segmentation strategy serves several purposes.
First, it reduces the chance of missing real behavioral boundaries.
If the first pass is too coarse, different behaviors can be merged before the semantic reasoning stage ever sees them.
Second, it preserves short transitions that would otherwise be absorbed into longer segments.
Third, it gives the VLM clips that are easier to interpret because each clip usually contains a smaller and more consistent set of actions.
Fourth, it leaves the final merging decision to the later LLM stage, where the model has access to richer semantic evidence from the generated descriptions.

DEC is also a low-cost first stage.
Training DEC on precomputed features is substantially cheaper than training a dedicated segmentation model such as an HMM-based pipeline end to end, and it is also lighter than workflows that first learn a separate representation with methods such as t-SNE-based pipelines and then perform clustering.
In practice, DEC provides a simple way to generate candidate segments without adding another expensive model-training step.

\begin{figure}
  \centering
  \includegraphics[width=\linewidth]{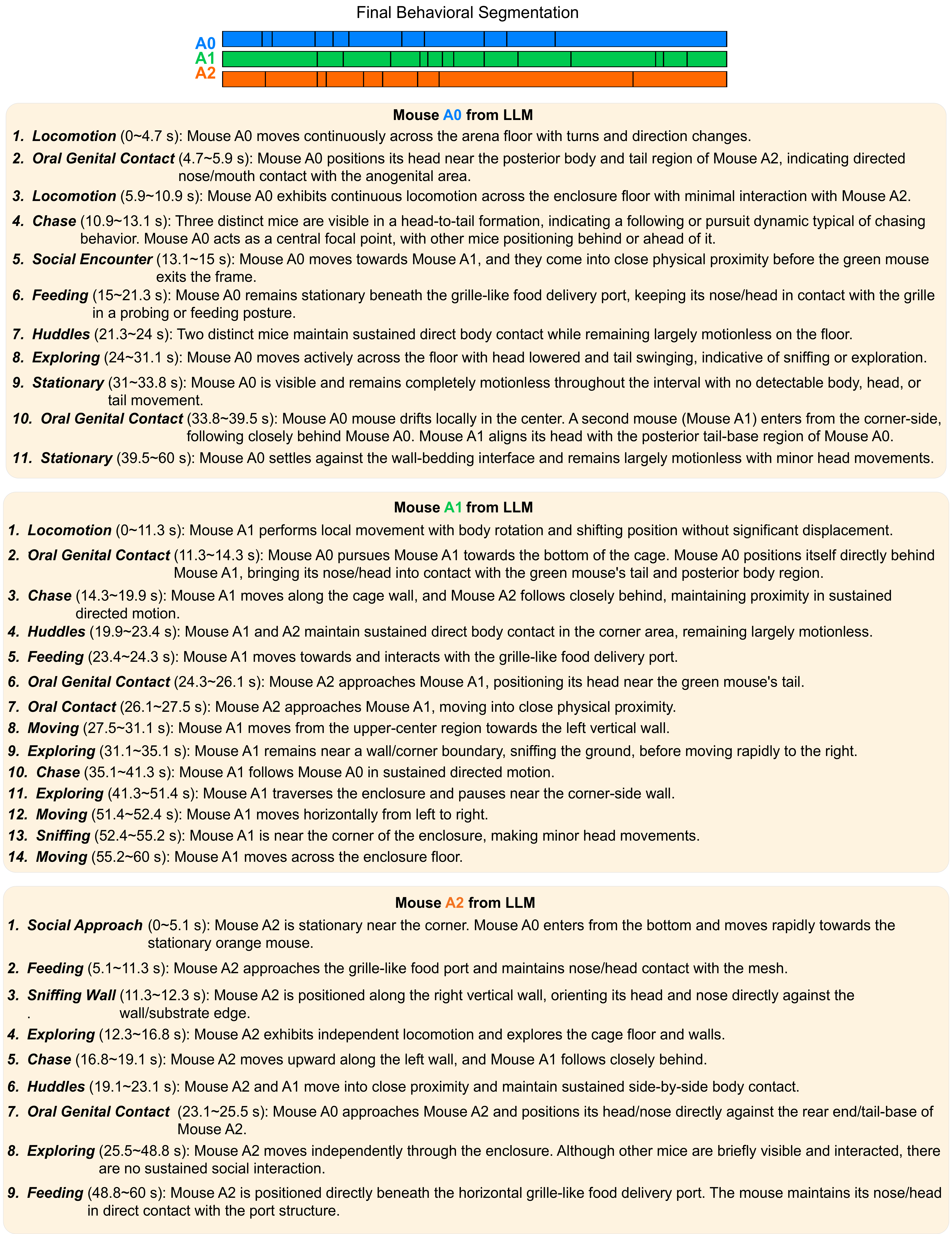}
  \caption{Behavioral understanding results for video 3ZOUFPHJ7JOHFBE8RHY6 in the MABe2022 Mouse Triplets dataset. BehaviorVLM produces temporally coherent behavioral segmentation for each mouse. For every candidate segment, a vision-language model (VLM) first generates natural-language descriptions of the observed actions and interactions (Figure~\ref{fig:behavior_pipeline}, mouse \texttt{A0} segments). A large language model (LLM) then refines and merges these descriptions into the final behavioral events shown here.}
  \label{fig:behavior_results}
\end{figure}

\paragraph{Stage 3: VLM-Based Per-Clip Video Understanding.}
For each short video clip produced by DEC, we invoke a state-of-the-art VLM to perform video understanding.
The VLM is prompted with a structured query that asks it to (i) assign a concise behavioral label, such as ``chasing'', ``exploring'', or ``feeding'', and (ii) generate a detailed natural-language description of the focal animal's behavior within the clip, including body posture, movement direction, speed, and any interactions with the other animals if they are present.
This stage converts each short clip into a textual representation without any task-specific training.
Although the initial segmentation is performed separately for each animal, the VLM still sees the video content of that animal's segment in its social context.
As a result, if the focal animal is interacting with another animal, the VLM can explicitly describe this interaction and produce social labels.
This is an important difference from traditional behavior segmentation pipelines: the segmentation is generated per animal, but the final labels can still contain social behavioral information because the VLM interprets the visual scene rather than only a single-animal motion trace.
The mouse \texttt{A0} segments shown in Figure~\ref{fig:behavior_pipeline} correspond to these direct VLM-stage segments and are intentionally more finely segmented than the final mouse \texttt{A0} segments shown in Figure~\ref{fig:behavior_results}.

\paragraph{Stage 4: LLM-Based Semantic Reasoning and Segment Merging.}
The set of per-clip text descriptions from the VLM is passed to an LLM with strong reasoning capability.
The LLM does not see the video directly.
Instead, the VLM serves as a perception module that converts visual observations into text, and the LLM serves as a reasoning module that organizes these textual descriptions into behaviors.
We use this separation deliberately because current state-of-the-art LLMs often provide stronger long-range semantic reasoning and grouping ability than state-of-the-art VLMs.
In this sense, the pipeline moves from perception to cognition: the VLM perceives the video and converts it into language, and the LLM performs higher-level reasoning over that perceived representation.
The LLM performs three operations.
First, it merges adjacent or nearby clips whose descriptions indicate the same behavioral state.
Second, it assigns a refined behavioral label and an enriched description to each merged segment by integrating evidence across the constituent clips.
Third, it returns a temporally structured behavioral annotation that can be used for downstream neuroscience analysis.
This is the stage that converts the finer VLM-stage segmentation into the longer final segmentation shown in Figure~\ref{fig:behavior_results}.

This pipeline requires no manually annotated behavior labels and no task-specific model training.
By combining the discriminative power of clustering in feature space with the semantic richness of VLMs and the reasoning capability of LLMs, BehaviorVLM achieves an interpretable and scalable solution to multi-animal behavioral understanding.
The method has several practical advantages.
It is guided by semantic understanding of behavior rather than only unstable low-level dynamics.
Its over-segmentation followed by semantic merging avoids committing too early to incorrect boundaries.
It combines video and text reasoning in a multimodal pipeline.
It groups behaviors by semantic meaning rather than only by pose or motion similarity.
It is robust to keypoint noise and can operate without keypoints.
It produces human-readable descriptions for each segment.
Overall, the pipeline mimics a human-like process in which visual observations are first described and then organized into coherent behaviors.

% We compare against Keypoint-MoSeq~\cite{weinreb2024keypoint}, trained with 10 latent dimensions and 10 shared states, as a representative unsupervised baseline, and assess segmentation quality and semantic interpretability qualitatively and quantitatively.

\subsection{Results}
\label{sec:behavior_results}

In our implementation, we use fused keypoint and visual features from~\citep{li2026learninglookondemandkeypointvideo} as inputs to the pipeline.
At the same time, the method is not restricted to keypoint-based representations.
It can also operate on visual features extracted directly from video, which is important because one goal of BehaviorVLM is to show that behavior segmentation can be performed from visual information alone rather than requiring keypoints.

For DEC-based clustering, we set the number of clusters to $K=10$ per animal, yielding short clips with average duration of approximately 1--5 seconds.
For VLM captioning, we use Qwen3.5-35B-A3B \citep{qwen3.5} with a structured prompt template, and all clips are uniformly downsampled to 10\,fps before being passed to the VLM.
Finally, the LLM reasoning step uses Qwen3-Next-80B-A3B \citep{qwen3technicalreport} with a prompt that receives clip descriptions for a contiguous behavioral epoch and outputs merged segments with refined labels.

\paragraph{Behavior Segmentation.}
Figure~\ref{fig:behavior_results} illustrates an example behavioral segmentation produced by BehaviorVLM on a multi-animal interaction sequence.
The model generates temporally coherent behavioral segments for each animal, shown in the segmentation timeline at the top of the figure.
Each segment corresponds to a contiguous interval with relatively consistent motion and interaction patterns.
BehaviorVLM produces boundaries that align well with visually identifiable behavioral transitions.
In contrast, purely kinematic unsupervised approaches often exhibit rapid state switching and fragmented segments because they rely on low-level motion statistics alone. The segmented videos corresponding to the final LLM-merged outputs for mouse A0 can be viewed at \url{https://tinyurl.com/video-for-segments-from-llm}.

\paragraph{Semantic Labels and Descriptions.}
Beyond segmentation, BehaviorVLM provides semantic annotations for each behavioral segment.
The VLM first describes short candidate clips, including the finer mouse \texttt{A0} segments shown in Figure~\ref{fig:behavior_pipeline}.
The LLM then merges these clips into the longer final segments shown in Figure~\ref{fig:behavior_results}.
This distinction matters because the VLM stage is intentionally more segmented, while the LLM stage is responsible for producing the final, easier-to-read behavioral timeline.
The system produces interpretable labels such as chasing, huddling, oral contact, and oral-genital contact, along with other behavior descriptions supported by the video evidence.
These segment-level explanations provide semantic information that is typically absent from behavior segmentation methods that return only latent states or cluster identities.

%-----------------------------------------------------------------------
\section{Conclusion}
\label{sec:conclusion}

We presented BehaviorVLM, a unified and finetuning-free framework for animal pose estimation and behavioral understanding.
For pose estimation, we introduced a multi-stage prompting pipeline that guides a pretrained vision-language model through sequential body-region detection, within-region and cross-region keypoint assignment, followed by RANSAC-based 3D cross-view consensus refinement.
Using only three manually annotated reference frames and no model fine-tuning, BehaviorVLM achieves reliable keypoint tracking across a 500-timepoint, six-view recording while producing labels that can be reviewed, filtered, corrected, and reused for downstream pose model training.
For behavioral understanding, we introduced a pipeline that combines low-cost deep embedded clustering for fine-grained candidate segments with vision-language models for clip-level interpretation and large language models for semantic refinement and segment merging.
This pipeline can use direct visual information and is not restricted to keypoint-based segmentation.
Together, these results highlight the promise of structured vision-language reasoning for neuroscience by reducing manual annotation burden while preserving interpretable intermediate outputs that researchers can inspect and reuse.

\bibliography{cited}
\bibliographystyle{plainnat}

\end{document}